\documentclass{article}

\usepackage{times}
\usepackage{graphicx} 
\usepackage{tikz}
\usetikzlibrary{calc}
\usetikzlibrary{positioning}
\usepackage{subfigure}

\usepackage{array}

\usepackage{color}
\usepackage{amsmath}

\usepackage{textcomp}
\usepackage{xspace}

\usepackage{placeins}

\usepackage{multirow}

\usepackage{natbib}

\usepackage{algorithm}
\usepackage{algorithmic}

\usepackage{hyperref}

\usepackage[accepted]{icml2017}

\newcommand{\optimizer}{{RNNprop}\xspace}
\newcommand{\textoptimizer}{{RNNprop}\xspace}

\newcommand{\lstmoptimizer}{{DMoptimizer}\xspace}
\newcommand{\textlstmoptimizer}{{DMoptimizer}\xspace}

\begin{document} 

\twocolumn[
\icmltitle{Learning Gradient Descent: Better Generalization and Longer Horizons}



\icmlsetsymbol{equal}{*}

\begin{icmlauthorlist}
\icmlauthor{Kaifeng Lv}{equal,iiis}
\icmlauthor{Shunhua Jiang}{equal,iiis}
\icmlauthor{Jian Li}{iiis}
\end{icmlauthorlist}

\icmlaffiliation{iiis}{Institute for Interdisciplinary Information Sciences, Tsinghua University, Beijing, China}

\icmlcorrespondingauthor{Kaifeng Lv}{vfleaking@163.com}
\icmlcorrespondingauthor{Shunhua Jiang}{linda6582@163.com}
\icmlcorrespondingauthor{Jian Li}{lijian83@mail.tsinghua.edu.cn}

\icmlkeywords{learning to learn, neural network, optimization}

\vskip 0.3in
]



\printAffiliationsAndNotice{\icmlEqualContribution \textsuperscript{\textdagger}The research is supported in part by the National Basic Research Program of China grants 2015CB358700, 2011CBA00300, 2011CBA00301, and the National NSFC grants 61632016.}


\begin{abstract} 
Training deep neural networks is a highly nontrivial task,
involving carefully selecting appropriate training algorithms, scheduling
step sizes and tuning other hyperparameters.
Trying different combinations can be quite labor-intensive and time consuming.
Recently, researchers have tried to use deep learning algorithms
to exploit the landscape of the loss function of the training problem of interest,
and learn how to optimize over it in an automatic way.
In this paper, we propose a new learning-to-learn model and some
useful and practical tricks. 
Our optimizer outperforms generic, hand-crafted optimization algorithms 
and state-of-the-art learning-to-learn optimizers by DeepMind in many tasks.
We demonstrate the effectiveness of our algorithms on a number of tasks, including deep MLPs, CNNs, and simple LSTMs.

\end{abstract} 

\section{Introduction}
Training a neural network can be viewed as solving an optimization problem 
for a highly non-convex loss function.
Gradient-based algorithms are by far the most widely used 
algorithms for training neural networks, such as
basic SGD, Adagrad, RMSprop, Adam, etc.  
For a particular neural network, 
it is unclear a priori which one is the best optimization algorithm,
and how to set up the hyperparameters (such as learning rates).
It usually takes a lot of time and experienced hands to identify 
the best optimization algorithm together with best hyperparameters,
and possibly some other tricks are 
necessary to make the network work.

\subsection{Existing Work} \label{sec:deepmind}

To address the above issue, a promising approach is to use machine learning 
algorithms to replace the hard-coded optimization algorithms, and hopefully,
the learning algorithm is capable of learning a good strategy, from experience,
to explore the landscape of the loss function and adaptively choose good descent steps.
In a high level, the idea can be categorized under the umbrella of {\em learning-to-learn} (or {\em meta-learning}), a broad area known to learning community for more than two decades.

Using deep learning for training deep neural networks was initiated  
in a recent paper \cite{Andrychowicz2016Learning}.
The authors proposed an optimizer using coordinatewise Long Short Term Memory
(LSTM) \cite{hochreiter1997long} that takes the gradients of the optimizee as input and outputs the updates for each optimizee parameters. We call this optimizer \lstmoptimizer throughout this paper, and we use the term \emph{optimizee} to refer to the loss function of the neural network being optimized. The authors showed that \lstmoptimizer outperforms traditional optimization algorithms in solving the task on which it is trained, and it also generalizes well to the same type of tasks. In one of their experiments, they trained \lstmoptimizer to minimize the average loss of a $100$-step training process of a $1$-hidden-layer Multilayer Perceptron (MLP) with sigmoid as the activation function, and the optimizer was shown to have generalization ability
to some extent: 
it also performs well on such MLP with one more hidden layer or double hidden neurons.
However, there are still some limitations:
\begin{enumerate}
\item 
If the activation function of the MLP is changed from sigmoid to ReLU in the test phase, \lstmoptimizer performs poorly to train such MLP.
In other words, their algorithms fail to generalize to different activations.

\item 
Even though the authors showed that \lstmoptimizer performs well to train the optimizee for $200$ descent steps, 
the loss increases dramatically for much longer horizons. In other words, their algorithms fail to handle a relatively large number of descent steps.
\end{enumerate}

\subsection{Our Contributions}

In this paper, we propose two new training tricks and a new model to improve the results of training a recurrent neural network (RNN) to optimize the loss functions of real-world neural networks.

The most effective trick is Random Scaling, which is used when training the RNN optimizer to improve its generalization ability by randomly scaling the parameters of the optimizee. The other trick is to combine the loss function of the optimizee with other simple convex functions, which helps to accelerate the training process. With the help of our new training tricks, our new model, called \optimizer, achieves notable improvements upon previous work after being trained on a simple $1$-hidden-layer MLP:
\begin{enumerate}
\item It can train optimizees for longer horizons. In particular, when \optimizer is only trained to minimize the final loss of a  $100$-step training process, in testing phase it can successfully train optimizees for several thousand steps.

\item It can generalize to a variety of neural networks including much deeper MLPs, CNNs, and simple LSTMs. On these tasks it achieves better or at least comparable performance with traditional optimization algorithms.
\end{enumerate}


\section{Other Related Work}
\subsection{Learning to Learn}

The notion of learning to learn or meta-learning has been used to address the concept of learning meta-knowledge about the learning process for years. However, there is no agreement on the exact definition of meta-learning, and various concepts have been developed by different authors \cite{Thrun1998Learning,Vilalta2002A,brazdil2008metalearning}.

In this paper, we view the training process of a neural network as an optimization problem, and we use an RNN as an optimizer to train other neural networks. The usage of another neural network to direct the training of neural networks has been put forward by Naik and Mammone \yrcite{Naik1992Meta}. In their early work, Cotter and Younger \yrcite{Cotter1990Fixed,Younger1999Fixed} argued that RNNs can be used to model adaptive optimization algorithms \cite{Prokhorov2002Adaptive}. This idea was further developed in \cite{Younger2001Meta,Hochreiter2001Learning} and gradient descent is used to train an RNN optimizer on convex problems. Recently, as shown in Section \ref{sec:deepmind}, Andrychowicz et al. \yrcite{Andrychowicz2016Learning} proposed a more general optimizer model using LSTM to learn gradient descent, and our work directly follows their work. In another recent paper \cite{Chen2016Learning}, an RNN is used to take current position and value as input and outputs the next position, and it works well for black-box optimization and simple RL tasks.

From a reinforcement learning perspective, the optimizer can be viewed as a policy which takes the current state as input and output the next action \cite{Schmidhuber1999Simple}. Two recent papers \cite{learning-step-size-controllers-for-robust-neural-network-training,Hansen2016Using} trained adaptive controllers to adjust the hyperparameters (learning rate) of traditional optimization algorithms from this perspective. Their method can be regarded as hyperparameter optimization. More general methods have been introduced in \cite{Li2016Learning, Wang2016Learning} which also take the RL perspective and train a neural network to model a policy.

\subsection{Traditional Optimization Algorithms}
A great number of optimization algorithms have been proposed to improve the performance of vanilla gradient descent, including Momentum\cite{tseng1998incremental},  Adagrad\cite{duchi2011adaptive}, Adadelta\cite{zeiler2012adadelta}, RMSprop\cite{tieleman2012lecture}, Adam\cite{kingma2014adam}. The update rules of several common optimization algorithms are listed in Table \ref{tab:algo-table}.

\begin{table}[htbp]
\caption{Traditional optimization algorithms. $\theta$ are the parameters of a neural network and $g$ represents the gradient. $\alpha, \beta_1, \beta_2, \gamma$ are the hyperparameters of an optimization algorithm. All vector operations are coordinatewise.}
\label{tab:algo-table}
\begin{center}
\begin{small}
\begin{tabular}{ll}
\hline
\abovespace\belowspace
Name & Update Rule \\
\hline
\abovespace
SGD & $\Delta \theta_t = -\alpha g_t$\\
\abovespace
Momentum & $m_{t} =  \gamma m_{t-1} + (1-\gamma)g_t$, \\
 &$\Delta \theta_t = -\alpha \, m_{t}$\\
\abovespace
Adagrad & $G_{t} =  G_{t-1} + g_t^2$, \\
 &$\Delta \theta_t= -\alpha \, g_t G_{t}^{-1/2}$\\
\abovespace
Adadelta & $v_{t} =  \beta_2 v_{t-1} + (1-\beta_2) g_t ^2$, \\
&$\Delta \theta_t =  -\alpha g_t v_t^{-1/2} D_{t-1}^{1/2}$, \\ 
&$D_t = \beta_1 D_{t-1} + (1-\beta_1) (\Delta \theta_{t} / \alpha)^2$\\
\abovespace
RMSprop & $v_{t} =  \beta_2 v_{t-1} + (1-\beta_2) g_t ^2$, \\ 
&$\Delta \theta_t= -\alpha \, g_t v_{t}^{-1/2}$\\
\abovespace
Adam & $m_{t} =  \beta_1 m_{t-1} + (1-\beta_1)g_t$, \\
&$v_{t} =  \beta_2 v_{t-1} + (1-\beta_2) g_t ^2$, \\
& $\hat{m}_{t} =  m_{t} /(1-\beta_1^t)$, \\
&$\hat{v}_{t} =  v_{t} / (1-\beta_2^t)$, \\
 &$\Delta \theta_t= -\alpha \, \hat{m}_t \hat{v}_{t}^{-1/2}$\belowspace\\
\hline
\end{tabular}
\end{small}
\end{center}
\vskip -0.05in
\end{table}

\section{Rethinking of Optimization Problems}
\subsection{Problem Formalization}
We are interested in finding an \textit{optimizer} that undertakes the optimization tasks for different optimizees. An \textit{optimizee} is a function $f(\theta)$ to be minimized. In the case when the optimizee is stochastic, that is, the value of $f(\theta)$ depends on the sample $d$ selected from a dataset $D$, the goal of an optimizer is to minimize
\begin{equation}
\frac{1}{\lvert D \rvert}\sum_{d \in D} f_d(\theta)
\end{equation}
over the variables $\theta$.

When optimizing an optimizee on a dataset $D$, the behavior of an optimizer
can be summarized by the following loop. For each step:
\begin{enumerate}
\item Given the current parameters $\theta_t$ and a sample $d_t \in D$, perform forward and backward propagation to compute the function value $y_t = f_{d_t}(\theta_t)$ and the gradient $g_t = \nabla f_{d_t}(\theta_t)$;
\item Based on the current state $\mathbf{h}_t$ 
(of the optimizer)
and the gradient $g_t$, the optimizer produces the new state $\mathbf{h}_{t+1}$ and proposes an increment $\Delta \theta_t$;
\item Update the parameters by setting $\theta_{t+1} = \theta_t + \Delta \theta_t$.
\end{enumerate}

In the initialization phase, $\mathbf{h}_0$ is produced by the optimizer, and $\theta_0$ is generated according to the initialization rule of the given optimizee. At the end of the loop, we take $\theta_T$ as the final optimizee parameters.

\subsection{Some Insight into Adaptivity}
Table \ref{tab:algo-table} summaries optimization algorithms that are most commonly used when training neural networks. All of these optimization algorithms have some degree of adaptivity, that is, they are able to adjust the effective step size $\lvert \Delta \theta_t \rvert$ when training.

We can divide these algorithms into two classes. The first class includes SGD and Momentum, as they determine the effective step size by the absolute size of gradients. The second class includes Adagrad, Adadelta, RMSprop, and Adam. 
These algorithms maintain the sum or the moving average of past gradients $g^2_t$, which can be seen as, with a little abuse of terminology, the second raw moment (or uncentered variance). Then, these algorithms produce the effective step size only by the relative size of the gradient, namely, the gradient divided by the square root of the second moment coordinatewise.

In a training process, as the parameters gradually approach to a local minimum, a smaller effective step size is required for a more careful local optimization.
To obtain such smaller effective step size, these two classes of algorithms have two different mechanisms. For the first class, if we take the full gradient, the effective step size automatically gets smaller when approaching to a local minimum. 
However, since we use stochastic gradient descent, the effective step size may not be small enough, even if $\theta$ is not far from a local minimum.
For the second class, a smaller effective step size $\lvert \Delta \theta_{t,i} \rvert$ of each coordinate $i$ is mainly induced by a relatively smaller partial derivative comparing with past partial derivatives. When approaching to a local minimum, the gradient may fluctuate due to stochastic nature.
Algorithms of the second class can decrease the effective step size of each coordinate in accordance with the fluctuation amplitude of that coordinate, i.e.,
a coordinate with larger uncentered variance yields smaller effective step size.
Thus, the algorithms of the second class are able to further decrease effective step size for the coordinates with more uncertainty, and they are more robust than those of the first class.


To get more insight into the difference between these two classes of algorithms, we consider what happens if we scale the optimizee by a factor $c$, i.e., let $\tilde{f}(\theta) = c f(\theta)$. 
Ideally, the scaling should not affect the behaviors of the algorithms. However, for the algorithms of the first class, since $\nabla \tilde{f}(\theta) = c\nabla f(\theta)$, the effective step size is also scaled by $c$. Hence, the behaviors of the algorithms change completely. But for the algorithms of the second class, they behave the same on $\tilde{f}(\theta)$ and $f(\theta)$ since the scale factor $c$ is canceled out. 
Thus the algorithms of the second class are more robust with respect
to scaling.



The above observation, albeit very simple, is a key  inspiration for our new model.
On the one hand, we use some training tricks so that our model can be exposed to functions with different scales at the training stage. On the other hand, we take relative gradients as input so that our optimizer belongs to the second class. In the following section, we introduce our training tricks and new model in details.

\section{Methods}
Our RNN optimizer operates coordinatewise on parameters $\theta$, which follows directly from \cite{Andrychowicz2016Learning}. The RNN optimizer handles the gradients coordinatewise and maintains hidden states for every coordinate respectively. The parameters of the RNN itself are shared between different coordinates. In this way, the RNN optimizer can train optimizees with any number of parameters.
\subsection{Random Scaling}
\label{random scaling}
We propose a training trick, called \emph{Random Scaling}, to prevent overfitting when training our model. Before introducing our ideas, consider what happens if we train an RNN optimizer to minimize $f(\theta) = \lambda \| \theta \|_2^2$ with initial parameter $\theta_0$. Clearly, $\theta_{t+1} = \theta_t -  \frac{1}{2\lambda} \nabla f(\theta_t)$ is the optimal policy since the lowest point can be reached in just one step. However, if the RNN optimizer learns to follow this rule exactly, testing this RNN optimizer on the same function with different $\lambda$ might produce a modest or even bad result.

The method to solve this issue is rather simple: We randomly pick a $\lambda$ for every iteration when training our RNN optimizer. Notice that we can also pick a random number to scale all the parameters to achieve the same goal. To further generalize this idea, we design our training trick, \emph{Random Scaling}, which coordinatewise randomly scales the parameters of the objective function in the training stage.

In more details, for each iteration of training the optimizer on a loss function $f(\theta)$ with initial parameter $\theta_0$, we first randomly pick a vector $\mathbf{c}$ of the same dimension as $\theta$, where each coordinate of $\mathbf{c}$ is sampled independently from a distribution $D_0$. Then, we train our model on a new optimizee
\begin{equation}
f_{\mathbf{c}}(\theta) = f(\mathbf{c} \theta)
\end{equation}
with initial parameter $\mathbf{c}^{-1} \theta_0$, where all the multiplication and inversion operations are performed coordinatewise. In this way, the RNN optimizer is forced to learn an adaptive policy to determine the best effective step size, rather than to learn the best effective step size itself of a particular task.



\subsection{Combination with Convex Functions} \label{sec:convex}
Now we introduce another training trick. It is clear that we should train our RNN optimizer on optimizees implemented with neural networks. However, due to non-convex and stochastic nature of neural networks, it may be hard for an RNN to learn the basic idea of gradient descent.

Our idea is loosely inspired by the proximal algorithms
(see e.g., \cite{parikh2014proximal}). To make training easier, we combine the original optimizee function $f$ with an $n$-dim convex function $g$ to get a new optimizee function $F$
\begin{equation}
F(\theta, \mathbf{x}) = f(\theta) + g(\mathbf{x}).
\end{equation}
For every iteration of training RNN optimizer, we generate a random vector $\mathbf{v}$ in $n$-dim vector space, and the function $g$ is defined as
\begin{equation}
\label{func:g}
g(\mathbf{x}) = \frac{1}{n} \sum_{i = 1}^{n} (x_i - v_i)^2,
\end{equation}
where the initial value of $\mathbf{x}$ is also generated randomly.

Without this trick, the RNN optimizer wanders around aimlessly on the non-convex loss surface of function $f$ in the beginning stage of training. 
After we combine the optimizee with function $g$, 
since $g$ has the good property of convexity, our RNN optimizer 
soon learns some basic knowledge of gradient descent from these additional optimizee coordinates. This knowledge is shared with other coordinates because the RNN optimizer processes its input coordinatewise. In this way, we can accelerate the training process of the RNN optimizer. As the training continues, 
the RNN optimizer further learns a better method with gradient decent as a baseline. 

We can apply Random Scaling on the function $g$ as well to make the behavior of the RNN optimizer more robust.

\subsection{\textoptimizer Model} \label{sec:model}

\begin{figure}[htb]
\begin{center}

\centering
\begin{tikzpicture}

\tikzset{
	layer/.style={draw,rectangle,thick,minimum height=0.7cm},
	filter/.style={circle, draw, thick, minimum size=7mm, inner sep=1pt, 
      path picture={%
           \draw[thick, rounded corners] 
              (path picture bounding box.center)--++(65:2mm)--++(0:1mm);
           \draw[thick, rounded corners] 
              (path picture bounding box.center)--++(245:2mm)--++(180:1mm);
      }},
	times/.style={draw, circle, thick, minimum size=3mm, inner sep=0pt,
	  path picture={%
           \draw[thick] (path picture bounding box.center)--++(0.7mm,0.7mm)--++(-1.4mm,-1.4mm);
           \draw[thick] (path picture bounding box.center)--++(-0.7mm,0.7mm)--++(1.4mm,-1.4mm);
      }},
};

\node (mt) at (0, 0cm) {$\tilde{m}_t$};

\node (gt) at (0.6cm, 0cm) {$\tilde{g}_t$};

\node[layer, minimum width=2.4cm] (p) at ($(gt)!0.5!(mt)+(0, -1.3cm)$) {Preprocessing};
\node[layer, minimum width=2cm, below=0.6cm of p] (r) {RNN};
\node[filter, right=0.7cm of r] (tanh) {};
\node[times, right=0.7cm of tanh] (times) {};
\node[above=0.7cm of times] (alpha) {$\alpha$};
\node[right=0.7cm of times] (o) {$\Delta \theta_t$};

\coordinate[above=0cm of p] (pu);
\coordinate[left=0cm of r] (rnnl);
\coordinate[right=0cm of r] (rnnr);
\coordinate (ind) at ($(mt)+(0, -0.3cm)$);
\draw[->, thick] (mt |- ind) -- (mt |- pu);
\draw[->, thick] (gt |- ind) -- (gt |- pu);
\draw[->, thick] (p) -- (r);
\draw[->, thick] (r) -- (tanh);
\draw[->, thick] (alpha) -- (times);
\draw[->, thick] (tanh) -- (times);
\draw[->, thick] (times) -- (o);

\draw[->, thick] ($(rnnl)+(0,-1.5mm)$) to [bend right=155, looseness=2] ($(rnnr)+(0,-1.5mm)$);

\end{tikzpicture}
\caption{The structure of our model \optimizer.
}
\label{fig:model}
\end{center}
\vskip -0.2in
\end{figure}
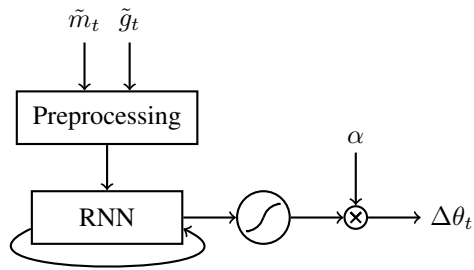

Aside from the above two tricks, we also design 
a new model \optimizer as shown in Figure \ref{fig:model}. All the operations in our model are coordinatewise, following the idea of \lstmoptimizer idea in \cite{Andrychowicz2016Learning}.

The main difference between \optimizer and \lstmoptimizer is the input. The input $\tilde{m}_t$ and $\tilde{g}_t$ are defined as follows:
\begin{eqnarray}
\tilde{m}_t &=& \hat{m}_t \hat{v}_t^{-1/2}, \\
\tilde{g}_t &=& g_t \hat{v}_t^{-1/2},
\end{eqnarray}
where $\hat{m}_t, \hat{v}_t$ are defined the same way as Adam in Table \ref{tab:algo-table}. 
This change of the input has three advantages. First, this input contains no information about the absolute size of gradients, so our algorithm belongs to the second class automatically and hence is more robust. Second, this manipulation of gradients can be seen as a kind of normalization so that the input values are bounded by a constant, which is somewhat easier for a neural network to learn. Lastly, if our model outputs a constant times $\tilde{m}_t$, it reduces to Adam. Similarly, if our model outputs a constant times $\tilde{g}_t$, then it reduces to RMSprop. 
Hence, the hope is that by further optimizing the parameters of \optimizer, it is capable of achieving better performance than Adam and RMSprop with fixed learning rate.

The input is preprocessed by a fully-connected layer with ELU (Exponential Linear Unit) as the activation function \cite{Clevert2015Fast} before being handled by the RNN. The central part of our model is the RNN, which is a two-layer coordinatewise LSTM that is same as \lstmoptimizer. The RNN outputs a single vector $\mathbf{x}_{\mathrm{out}}$, and the increment is taken as
\begin{equation}
\Delta \theta_t = \alpha \, \tanh(\mathbf{x}_{\mathrm{out}}).
\end{equation}
This formula can be viewed as a variation of gradient clipping so that all effective step sizes are bounded by the preset parameter $\alpha$. 
In all our experiments, we just set a large enough value $\alpha = 0.1$.

\begin{figure*}[tb]
\begin{center}
\centerline{\includegraphics[width=2\columnwidth]{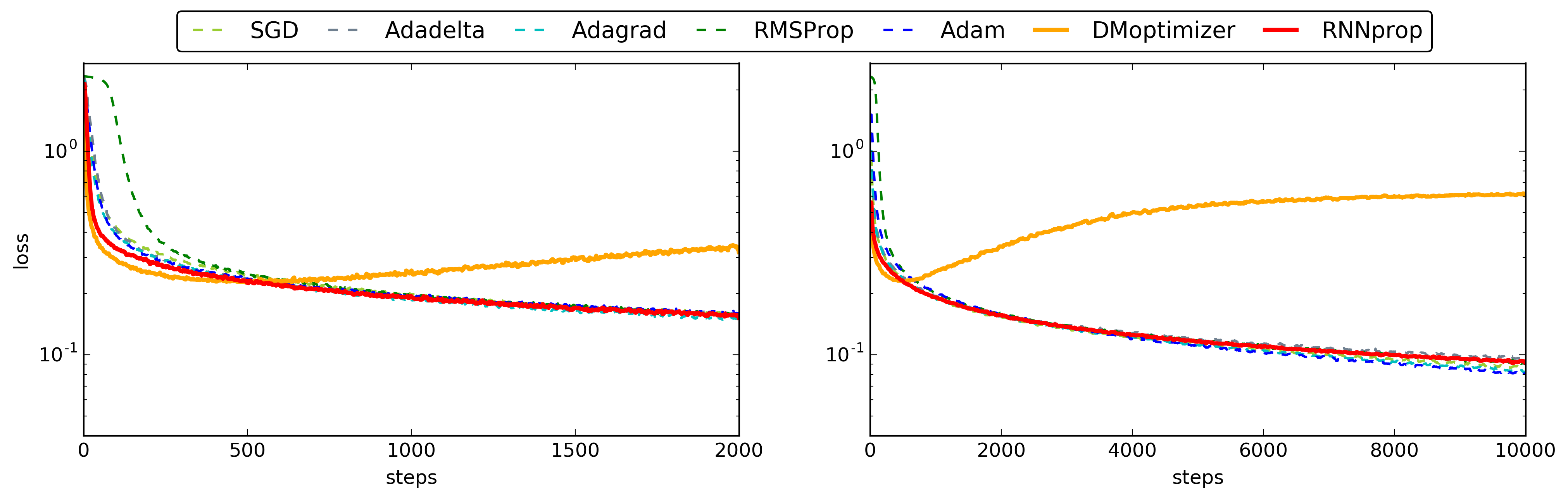}}
\caption{Performance on the base MLP. \textbf{Left:} \optimizer achieves comparable performance when allowed to run for $2000$ steps. \textbf{Right:} \optimizer continues to decrease the loss even for $10000$ steps, but the performance is slightly worse than some traditional algorithms.}
\label{fig:longer-horizon}
\end{center}
\vskip -0.25in
\end{figure*}

\section{Experiments}

We trained two RNN optimizers, one to reproduce \lstmoptimizer in \cite{Andrychowicz2016Learning}, the other to implement \optimizer with our new training tricks. Their performances were compared in a number of experiments. \footnote{Our code can be found at \url{https://github.com/vfleaking/rnnprop}.}

We use the same optimizee as in \cite{Andrychowicz2016Learning} to train these two optimizers, which is the cross-entropy loss of a simple MLP on the MNIST dataset. For convenience, we address this MLP as the base MLP. It has one hidden layer of $20$ hidden units and uses sigmoid as activation function. The value of $f(\theta)$ is computed using a minibatch of $128$ random pictures. For each iteration during training, the optimizers are allowed to run for $100$ steps. Optimizers are trained using truncated Backpropagation Trough Time (BPTT). We split the $100$ steps into $5$ periods of $20$ steps. In each period, we initialize the initial parameter $\theta_0$ and initial hidden state $\mathbf{h}_0$ from the last period or generate them if it is the first period. Adam is used to minimize the loss
$
\mathcal{L}(\phi) = \frac{1}{T} \sum_{t = 1}^{T} w_t f(\theta_t).
$
We trained \lstmoptimizer using the loss with $w_t = 1$ for all $t$ as in \cite{Andrychowicz2016Learning}. For \optimizer we set $w_T = 1$ and $w_t = 0$ for other $t$. In this way, the optimizer is not strictly required to produce a low loss at each step, so it can be more flexible. We also notice that this loss results in slightly better performance.

The structure of our model \optimizer is shown in Section \ref{sec:model}. The RNN is a two-layer LSTM whose hidden state size is $20$. To avoid division by zero, in actual experiments we add another term $\epsilon = 10^{-8}$, and the input is changed to
\begin{eqnarray}
\tilde{m}_t &=& \hat{m}_t (\hat{v}_t^{1/2} + \epsilon)^{-1}, \\
\tilde{g}_t &=& g_t (\hat{v}_t^{1/2} + \epsilon)^{-1}.
\end{eqnarray}
The parameters $\beta_1$ and $\beta_2$ for computing $m_t$ and $g_t$ are simply set to $0.95$. 
In preprocessing, the input is mapped to a $20$-dim vector for each coordinate.

When training \optimizer, we first apply Random Scaling to the optimizee function $f$ and the convex function $g$ respectively, where $g$ is defined as Equation \eqref{func:g}, and then we combine them together as introduced in Section \ref{sec:convex}. We set the dimension of the convex function $g$ to be $n = 20$ and generate the vectors $\mathbf{v}$ and $\mathbf{x}$ from $[-1, 1]^n$ uniformly randomly. To generate each coordinate of the vector $\mathbf{c}$ in Random Scaling, we first generate a number $p$ from $[-L, L]$ uniformly randomly, and then take $\exp(p)$ as the value of that coordinate, where $\exp$ is the natural exponential function. This implementation is aimed to produce $\mathbf{c}$ of different order of magnitude, e.g., $\Pr[\frac{1}{10} \leq c_i \leq \frac{1}{9}] = \Pr[9 \leq c_i \leq 10]$. We also tried other transformations including using uniform distribution, scaling the entire function directly, randomly dropping some coordinates, etc. This version of Random Scaling is selected after comprehensive comparison. In the experiments we set $L = 3$ for the function $f$ and $L = 1$ for the function $g$. 

We save all the parameters of the RNN optimizers every $1000$ iterations when training. For \lstmoptimizer, we select the saved optimizer with the best performance on the validation task, same as in \cite{Andrychowicz2016Learning}. Since \optimizer tends not to overfit to the training task because of the Random Scaling method, we simply select the saved optimizer with lowest average train loss, which is the moving average of the losses of the past $1000$ iterations with decay factor $0.9$. The selected optimizers are then tested on other different tasks. Their performances are compared with the best traditional optimization algorithms whose learning rates are carefully chosen and other hyperparameters are set to the default values in Tensorflow \cite{Abadi2016TensorFlow}. All the initial optimizee parameters used in the experiments are generated independently from the Gaussian distribution $N(0,0.1)$. 

All figures shown in this section were plotted after running the optimization process multiple times with random initial values and data. We removed the outliers with exceedingly large loss value when plotting the loss curves. No loss value of \optimizer was removed when plotting the figures.


\subsection{Generalization to More Steps}

We first test optimizers on the task used in the training stage, which is to optimize the base MLP for $100$ steps. Both \lstmoptimizer and \optimizer outperform all traditional optimization algorithms. \lstmoptimizer has better performance possibly because of overfitting. We then test optimizers to run for more steps on the base MLP. The left plot of Figure \ref{fig:longer-horizon} indicates that \optimizer can achieve comparable performance with traditional algorithms for $2000$ steps while \lstmoptimizer fails.

We also test the optimizers for much more steps: $10000$ steps, as shown in the right plot of Figure \ref{fig:longer-horizon}. It is clear that \lstmoptimizer loses the ability to decrease the loss after about $400$ steps and its loss begins to increase dramatically. \optimizer, on the other hand, is able to decrease the loss continuously, though it slows down gradually and traditional algorithms overtake it. The main reason is that \optimizer is trained to run for only $100$ steps, and $10000$-step training process may be significantly different from $100$-step training process. Additionally, traditional optimization algorithms are able to achieve good performance on both tasks because we explicitly adjusted their learning rates to adapt to these tasks.

\begin{figure}[htb]
\begin{center}
\centerline{\includegraphics[width=\columnwidth]{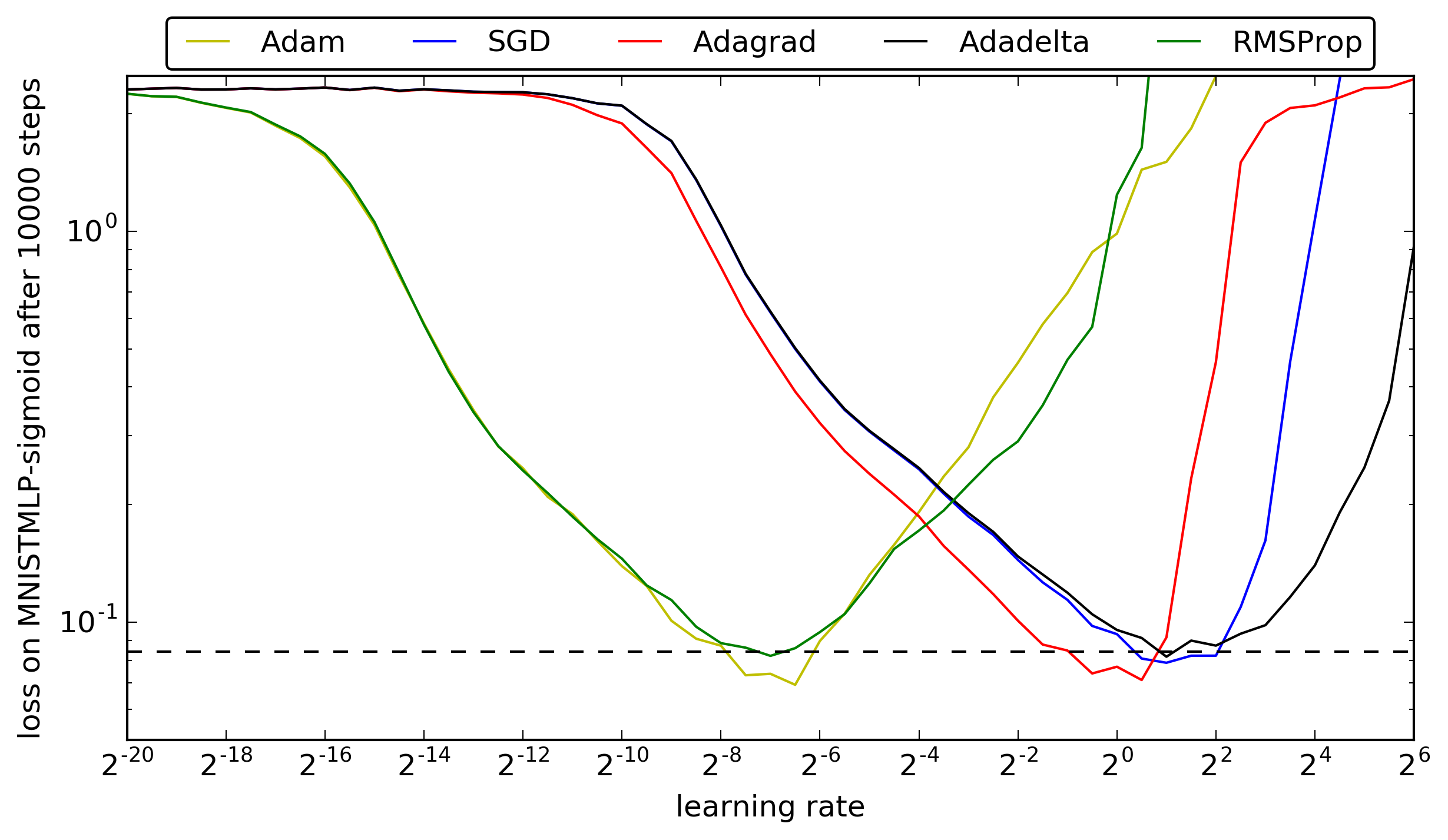}}
\caption{The final loss of different algorithms on the base MLP after $10000$ steps. The colorful solid curves show how the final losses of traditional algorithms after $10000$ steps change with different learning rates, and the horizontal dash line shows the final loss of \optimizer. We compute the final loss by freezing the final parameters of the optimizee and compute the average loss using all the data encountered during optimization process.}
\label{fig:lr10000}
\end{center}
\vskip -0.2in
\end{figure}

Figure \ref{fig:lr10000} shows how the final loss after $10000$ steps changes when using different learning rates. For example, Adam can outperform \optimizer only if its learning rate lies in the narrow interval from $0.004$ to $0.01$.

For other optimizees, \optimizer shows similar ability to train for longer horizons. Due to space constraints, we do not discuss them in details.

\subsection{Generalization to Different Activation Functions}
\begin{figure}[htb]
\begin{center}
\centerline{\includegraphics[width=\columnwidth]{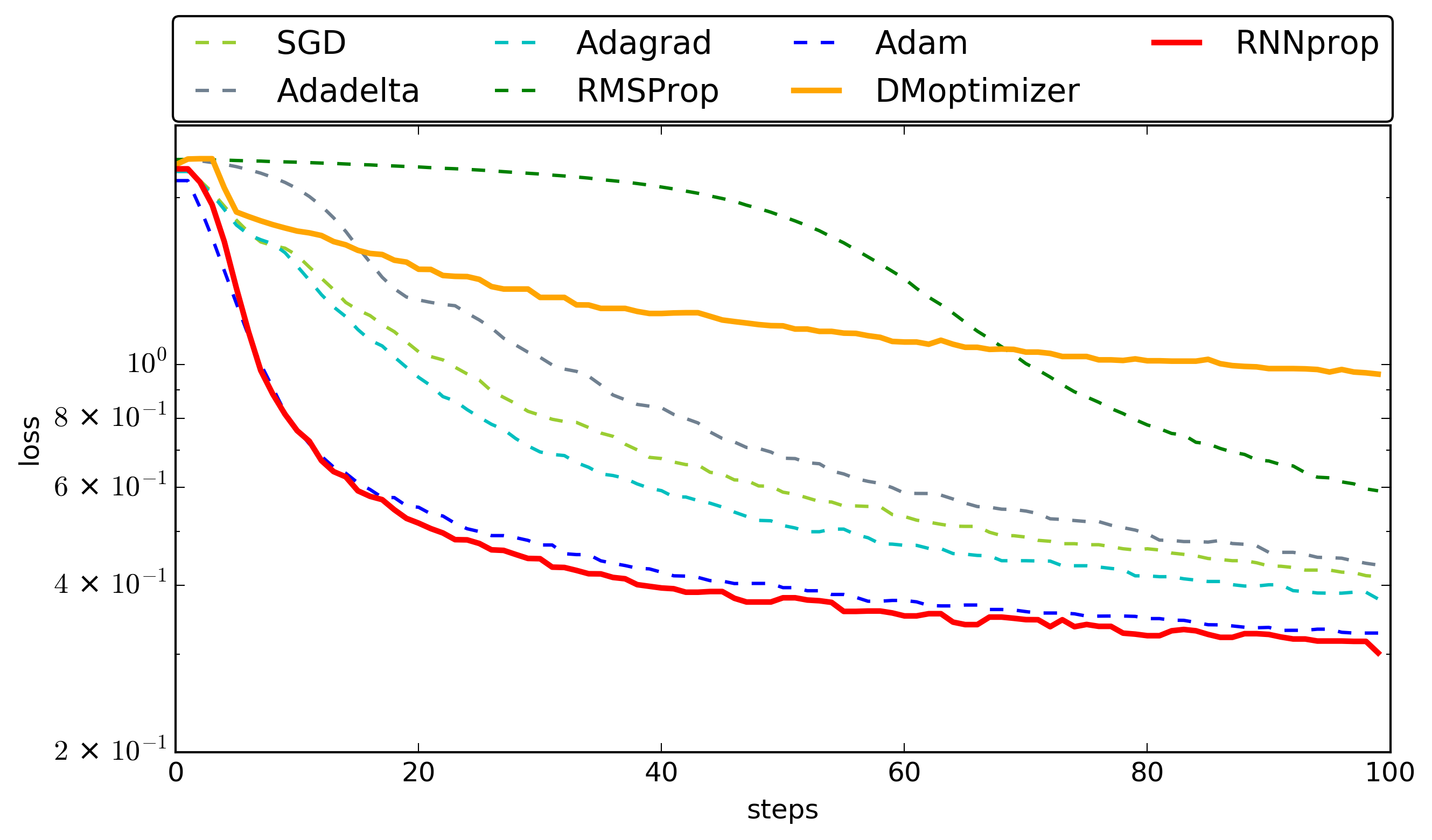}}
\caption{\optimizer slightly outperforms traditional algorithms on the base MLP with activation replaced with ReLU.}
\label{fig:relu}
\end{center}
\vskip -0.1in
\end{figure}

We test the optimizers on the base MLP with different activation functions. As shown in Figure \ref{fig:relu}, if the activation function is changed from sigmoid to ReLU, \optimizer can still achieve better performance than traditional algorithms while \lstmoptimizer fails. For other activations, \optimizer also generalizes well as shown in Table \ref{tab:act-mlp}.

\begin{table}[htb]
\caption{Performance on the base MLP with different activations. The numbers in table were computed after running the optimization processes for 100 times.}
\label{tab:act-mlp}
\begin{center}
\begin{small}
\begin{tabular}{c|c|c|c}
\hline
\abovespace\belowspace
Activation & Adam & \textlstmoptimizer & \textoptimizer \\
\hline
\abovespace
sigmoid  & $0.31$ & $\mathbf{0.24}$ & $0.29$ \\
\abovespace
ReLU  & $0.28$ & $1.05$ & $\mathbf{0.27}$ \\
\abovespace
ELU  & $0.26$ & $13.51$  & $\mathbf{0.24}$ \\
\abovespace
tanh  & $0.31$ & $0.50$    & $\mathbf{0.28}$ \belowspace\\
\hline
\end{tabular}
\end{small}
\end{center}
\end{table}

\subsection{Generalization to Deeper MLP}
In deep neural networks, different layers may have different optimal learning rates, but traditional algorithms only have one global learning rate for all the parameters. Our RNN optimizer can achieve better performance benefited from its more adaptive behavior.

\begin{figure}[tb]
\begin{center}
\centerline{\includegraphics[width=\columnwidth]{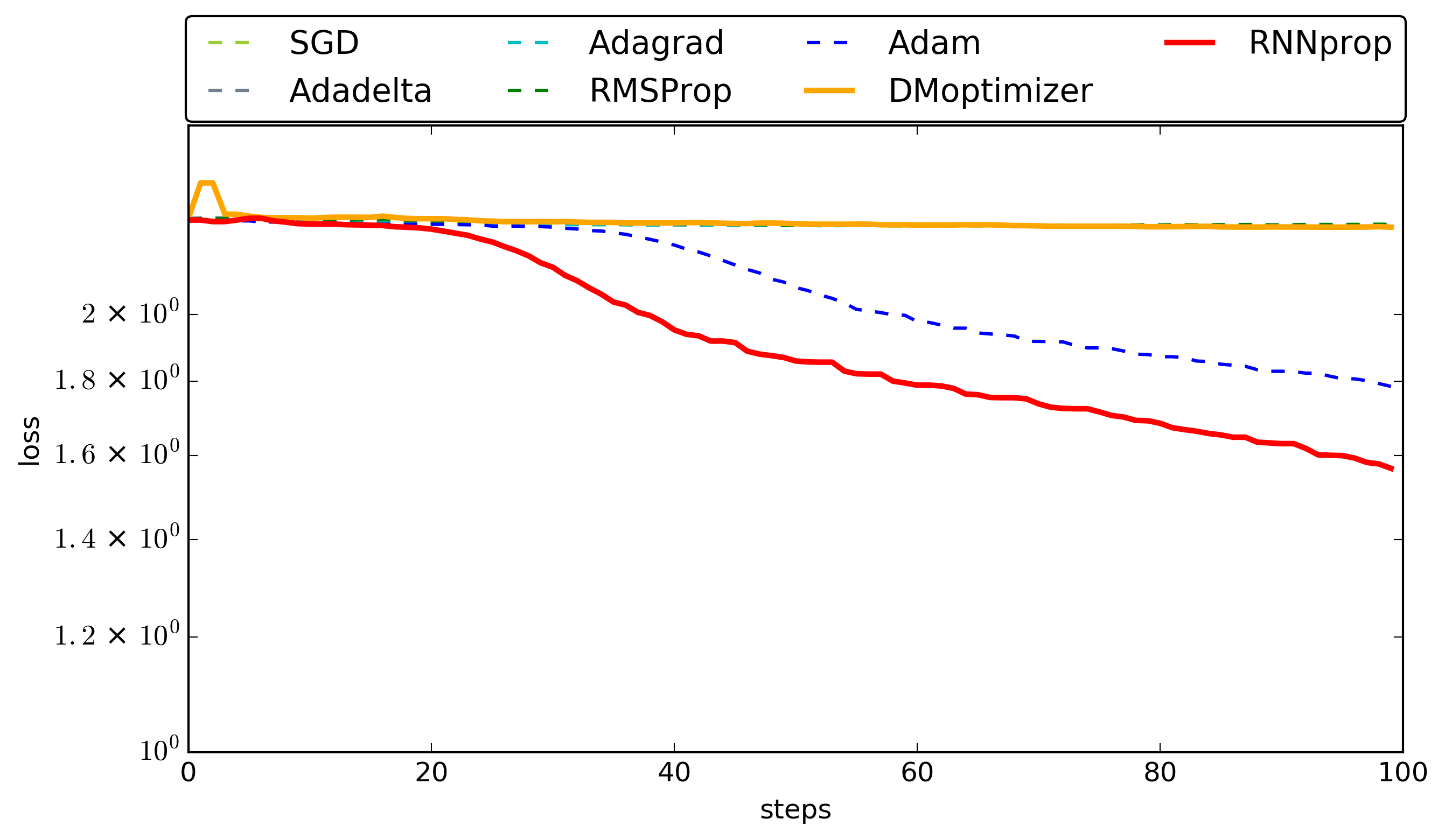}}
\caption{\optimizer significantly outperforms traditional algorithms on the base MLP with $5$ hidden layers.}
\label{fig:l5}
\end{center}
\vskip -0.1in
\end{figure}

\begin{figure}[tb]
\begin{center}
\centerline{\includegraphics[width=\columnwidth]{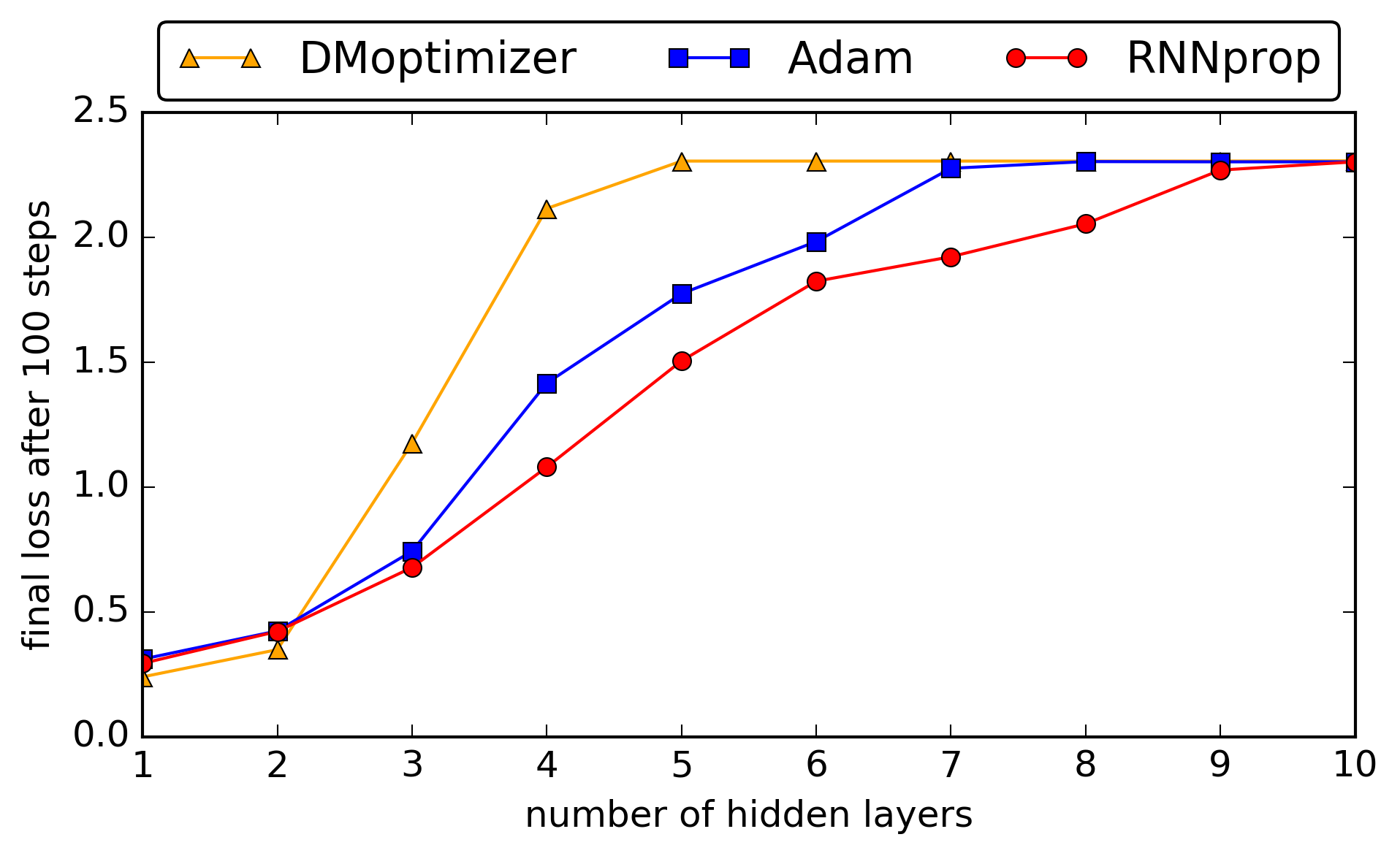}}
\caption{Performance on the base MLP with different number of hidden layers. Among all traditional algorithms we only list the performance of Adam since it achieves lowest loss.}
\label{fig:layer}
\end{center}
\vskip -0.2in
\end{figure}

We tested the optimizers on deeper MLPs. More hidden layers are added to the base MLP, all of which have $20$ hidden units and use sigmoid as activation function. As shown in Figure \ref{fig:layer}, \optimizer can always outstrip traditional algorithms until the MLP becomes too deep and none of them can decrease its loss in $100$ steps. Figure \ref{fig:l5} shows the loss curves on the MLP with $5$ hidden layers as an example.


\subsection{Generalization to Different Structures}
\subsubsection{CNN}
The CNN optimizees are the cross-entropy losses of convolutional neural networks (CNN) with similar structure as VGGNet \cite{simonyan2014very} on dataset MNIST or dataset CIFAR-10. All convolutional layers use $3\times 3$ filters and the window of each max-pooling layer is of size $2\times 2$ with stride $2$. We use \texttt{c} to denote a convolutional layer, \texttt{p} to denote a max-pooling layer and \texttt{f} to denote a fully-connected layer. Three CNNs are used in the experiments: CNN with structure \texttt{c-c-p-f} on MNIST, CNN with structure \texttt{c-c-p-c-c-p-f-f} on MNIST and CNN with structure \texttt{c-c-p-f} on CIFAR-10.

\begin{figure*}[t]
\begin{center}
\centerline{\includegraphics[width=2\columnwidth]{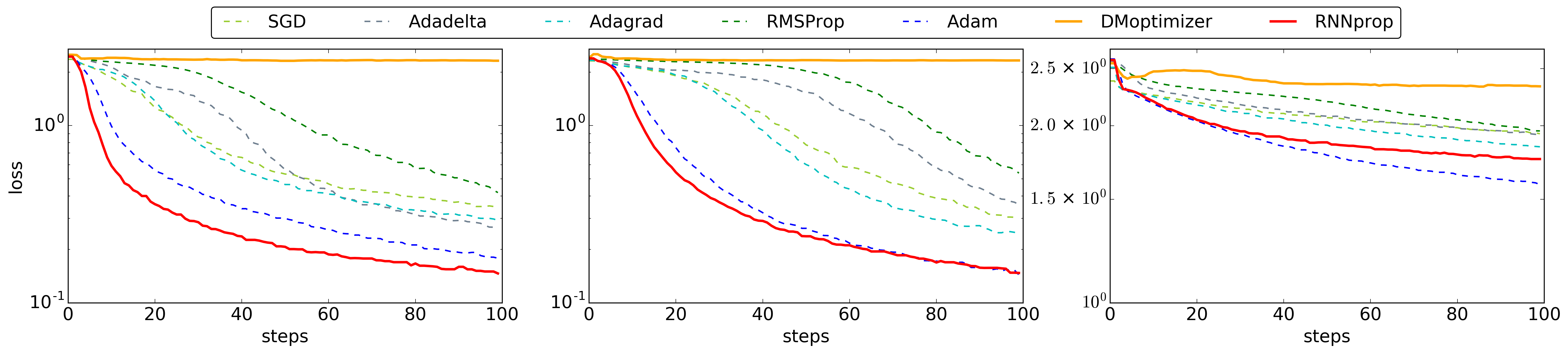}}
\caption{Performance on different CNNs. \textbf{Left: }The CNN has 2 convolutional layer, 1 pooling layer and 1 fully-connected layer and is on dataset MNIST. \textbf{Center: }The CNN has 4 convolutional layer, 2 pooling layer and 2 fully-connected layer and is on dataset MNIST. \textbf{Right: }The CNN has 2 convolutional layer, 1 pooling layer and 1 fully-connected layer and is on dataset CIFAR-10.}
\label{fig:cnn}
\end{center}
\end{figure*}

The results are shown in Figure \ref{fig:cnn}. \optimizer can outperform traditional algorithms on CNN with structure \texttt{c-c-p-f} on dataset MNIST. On the other two CNNs, only the best traditional algorithm outperforms \optimizer. Even though \cite{Andrychowicz2016Learning} showed that \lstmoptimizer that is trained on CNNs can train CNNs faster than traditional algorithms, in our experiments \lstmoptimizer fails to train any of the CNNs when the training set is fixed to the base MLP.


\subsubsection{LSTM}


The optimizers are also tested on the mean squared loss of an LSTM with hidden state size $20$ on a simple task: given a sequence $f(0),\dots,f(9)$ with additive noise, the LSTM needs to predict the value of $f(10)$. Here $f(x) = A \sin(\omega x + \phi)$. When generating the dataset, we uniformly randomly choose $A \sim U(0, 10), \omega \sim U(0, \pi/2), \phi \sim U(0, 2\pi)$, and we draw the noise from the Gaussian distribution $N(0,0.1)$.

Even though the task is completely different from the task that is used for training, \optimizer still has comparable or even better performance than traditional algorithms, which may be due to the fact the structure inside LSTM is similar to that of the base MLP with sigmoid in between.

We also adjust the settings of the task. As shown in Figure \ref{tab:lstm}, \optimizer still achieve good results when we use a smaller noise from the distribution $N(0, 0.01)$ or use a two-layer LSTM instead of one-layer.


\begin{table}[tb]
\caption{Performance on the task with different settings of LSTM. We list the final loss of \optimizer and best traditional optimization algorithms on the task with $2$-layer LSTM and on the task with smaller noise. The numbers in table were computed after running the optimization processes for 100 times. 
}
\label{tab:lstm}
\vskip 0.15in
\begin{center}
\begin{small}
\setlength{\tabcolsep}{5pt}
\begin{tabular}{c|c|c|c|c}
\hline
\abovespace\belowspace
Experiment & Adam &Adagrad& \textlstmoptimizer & \textoptimizer \\
\hline
\abovespace
Default  & $0.62$ &$\mathbf{0.54}$& $26.43$ & $0.55$ \\
\abovespace
$2$ Layers  & $0.44$ &$0.65$& $5.06$ & $\mathbf{0.28}$ \\
\abovespace
Small Noise  & $0.39$ &$0.50$& $22.04$  & $\mathbf{0.36}$\belowspace\\
\hline
\end{tabular}
\end{small}
\end{center}
\end{table}

\begin{figure}[tbp]
\begin{center}
\centerline{\includegraphics[width=\columnwidth]{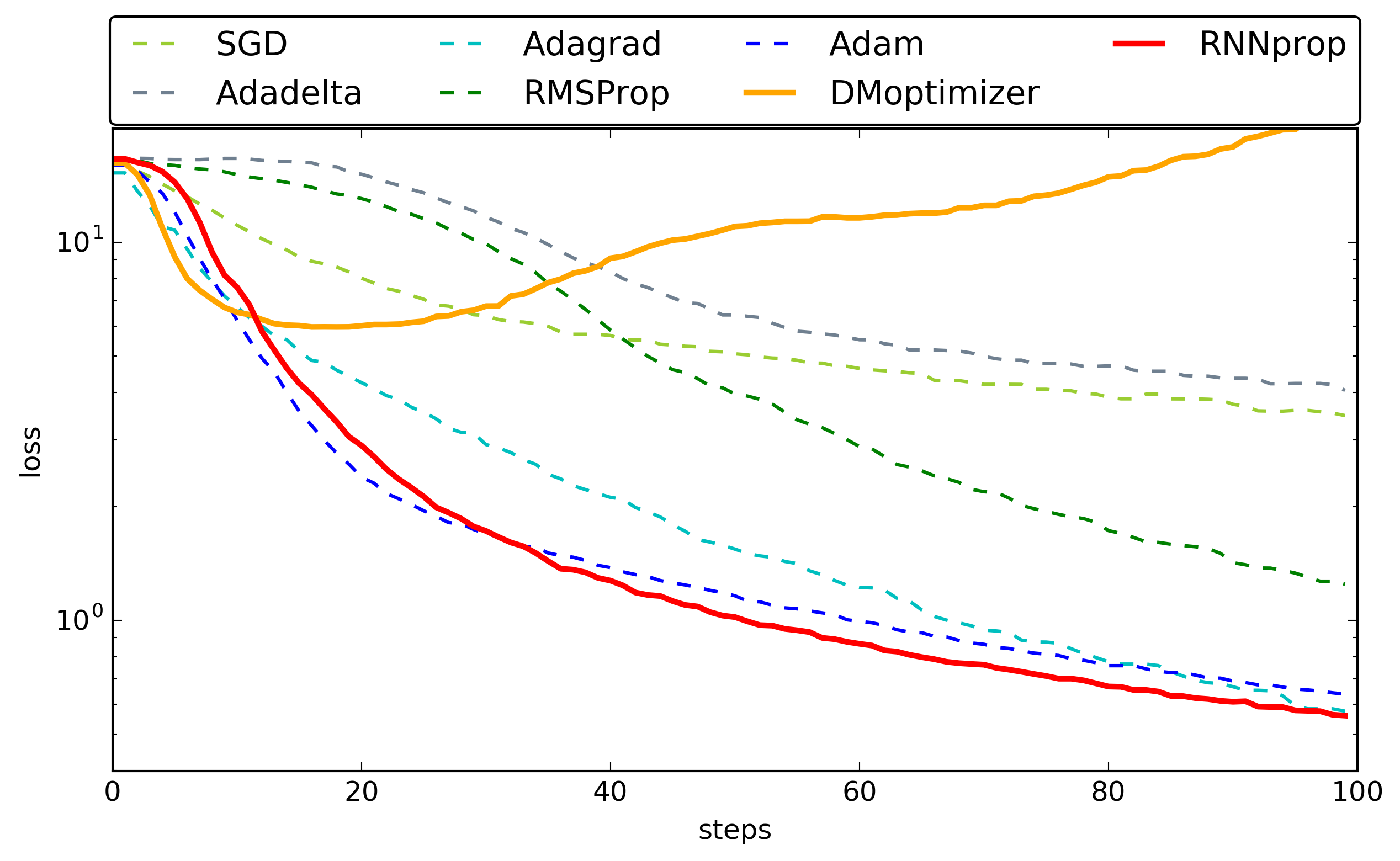}}
\caption{Performance on a sequence prediction problem implemented by LSTM.}
\label{fig:lstm}
\end{center}
\end{figure}

\subsection{Control Experiment}
To assess the effectiveness of each contribution separately, we also trained three more RNN optimizers: \lstmoptimizer trained with the two tricks and two \optimizer, each trained with one of the two tricks respectively. 

Recall that the trick of combining with convex function aims to accelerate the training of RNN optimizers. We test the performance of \optimizer whose own parameters are trained for different numbers of iterations, with or without this trick. The result is shown in Table \ref{tab:compare-mixed}. With this trick RNN optimizer can achieve a good result with fewer iterations of training.

\begin{figure}[tbp]
\begin{center}
\centerline{\includegraphics[width=\columnwidth]{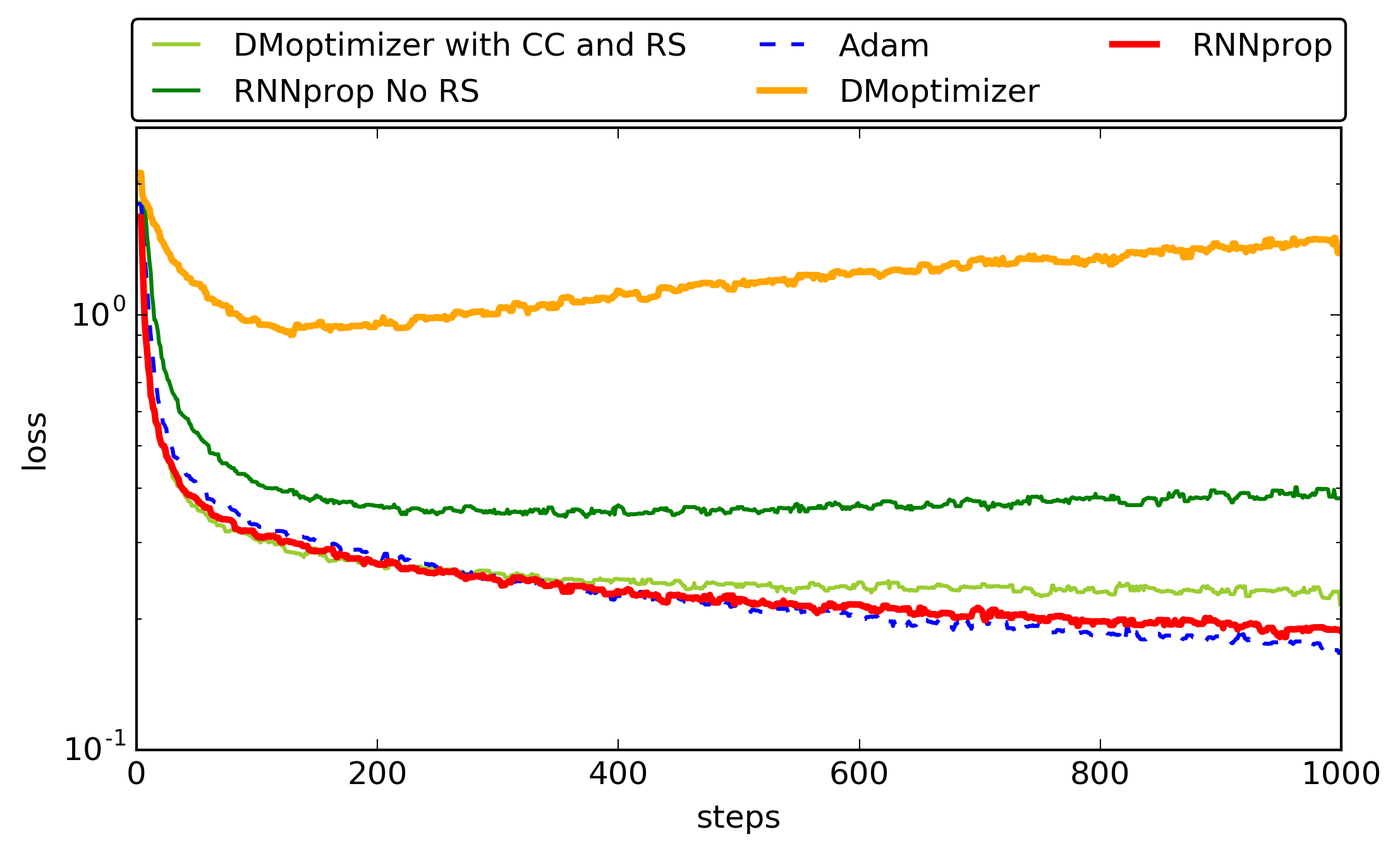}}
\caption{Comparison between RNNprop trained with two tricks, RNNprop trained without Random Scaling, DMoptimizer, DMoptimizer trained with two tricks. All optimizers are tested on the base MLP with activation replaced with ReLU for 1000 steps.}
\label{fig:comparison}
\end{center}
\vskip -0.1in
\end{figure}

To assess the other two contributions, we select the trained optimizers in the same way as \optimizer. In Figure \ref{fig:comparison}, we test their performances on the base MLP with activation replaced with ReLU for $1000$ steps. From the figure, we conclude that Random Scaling is the most effective trick.

\begin{table}[tbp]
\caption{Comparison between RNNprop with or without combination with convex functions. We test them on the base MLP and the base MLP with activation replaced with ReLU. The 2nd column shows the number of iterations used to train optimizers. The 4th column shows the final loss produced by RNNprop with all training tricks, while the last column shows the final loss produced by RNNprop trained without combination with convex functions.}
\label{tab:compare-mixed}
\begin{center}
\begin{small}
\begin{tabular}{c|c|c|c|c}
\hline
\abovespace\belowspace
Optimizee                 & \#Iter        & Adam   &  \textoptimizer & No CC \\
\hline
\abovespace
\multirow{3}{*}{Base MLP} & 5k        & $0.31$ & $\mathbf{0.33}$ & $\mathbf{0.33}$ \\
                          & 10k       & $0.31$ & $\mathbf{0.31}$ & $0.32$ \\
\belowspace
                          & 15k       & $0.31$ & $\mathbf{0.30}$ & $0.33$    \\
\hline
\abovespace
\multirow{3}{*}{ReLU}     & 5k        & $0.28$ & $\mathbf{0.30}$ & $0.31$ \\
                          & 10k       & $0.28$ & $\mathbf{0.29}$ & $0.31$ \\
\belowspace
                          & 15k       & $0.28$ & $\mathbf{0.27}$ & $0.32$    \\
\hline
\end{tabular}
\end{small}
\end{center}
\vskip -0.9in
\end{table}

\section{Conclusion}
In this paper, we present a new learning-to-learn model with several useful tricks. We show that our new optimizer has better generalization ability than the state-of-art learning-to-learn optimizers. After trained using a simple MLP, our new optimizer achieves better or comparable performance with traditional optimization algorithms when training more complex neural networks or when training for longer horizons.

We believe it is possible to further improve the generalization ability of our optimizer. Indeed, on some tasks in our experiments, our optimizer did not outperform the best traditional optimization algorithms, in particular when training for much longer horizon or when training neural networks on different datasets. 
In the future, we aim to further develop a more generic optimizer with more elaborate designing, so that it can achieve better performance on a wider range of tasks that are analogous with the optimizee used in training.


\FloatBarrier
\newpage
\bibliography{paper}
\bibliographystyle{icml2017}

\end{document}